\documentclass[letterpaper]{article} 
\usepackage{aaai2026}  
\usepackage{times}  
\usepackage{helvet}  
\usepackage{courier}  
\usepackage[hyphens]{url}  
\usepackage{graphicx} 

\usepackage{times}  
\usepackage{helvet}  
\usepackage{courier}  
\usepackage[hyphens]{url}  
\usepackage{graphicx} 
\usepackage{natbib}  
\usepackage{caption} 
\usepackage{booktabs}
\usepackage{multirow}

\usepackage{diagbox}

\usepackage{booktabs}   
\usepackage{multirow}   

\usepackage{algorithm}
\usepackage{algorithmic}
\usepackage{amsmath}

\usepackage{newfloat}
\usepackage{listings}

\usepackage{amsmath}

\usepackage{natbib}  
\usepackage{caption} 
\usepackage{algorithm}
\usepackage{algorithmic}

\usepackage{newfloat}
\usepackage{listings}
\DeclareCaptionStyle{ruled}{labelfont=normalfont,labelsep=colon,strut=off} 
\floatstyle{ruled}
\newfloat{listing}{tb}{lst}{}
\floatname{listing}{Listing}
\title{Semore: VLM-guided Enhanced Semantic Motion Representations for Visual Reinforcement Learning}
\author{
    Wentao Wang\textsuperscript{\rm 1},
    Chunyang Liu\textsuperscript{\rm 2},
    Kehua Sheng\textsuperscript{\rm 2},
    Bo Zhang\textsuperscript{\rm 2}\equalcontrib,
    Yan Wang\textsuperscript{\rm 1}\equalcontrib
}
\affiliations{
    \textsuperscript{\rm 1} Institute for AI Industry Research, Tsinghua University\\

   \textsuperscript{\rm 2} Didi Chuxing\\

   kensorpl@aliyun.com, \{liuchunyang, shengkehua, zhangbo\}@didiglobal.com,  wangyan@air.tsinghua.edu.cn
    
}

\usepackage{bibentry}

\begin{document}

\maketitle

\begin{abstract}
The growing exploration of Large Language Models (LLM) and Vision-Language Models (VLM) has opened avenues for enhancing the effectiveness of reinforcement learning (RL). However, existing LLM-based RL methods often focus on the guidance of control policy and encounter the challenge of limited representations of the backbone networks. To tackle this problem, we introduce Enhanced Semantic Motion Representations (Semore), a new VLM-based framework for visual RL, which can simultaneously extract semantic and motion representations through a dual-path backbone from the RGB flows. Semore utilizes VLM with common-sense knowledge to retrieve key information from observations, while using the pre-trained clip to achieve the text-image alignment, thereby embedding the ground-truth representations into the backbone. To efficiently fuse semantic and motion representations for decision-making, our method adopts a separately supervised approach to simultaneously guide the extraction of semantics and motion, while allowing them to interact spontaneously. Extensive experiments demonstrate that, under the guidance of VLM at the feature level, our method exhibits efficient and adaptive ability compared to state-of-art methods. All codes are released ${}^1$.
\end{abstract}


\section{Introduction}

Thanks to its ability to directly convert complex visual signals into actions, visual Reinforcement Learning (RL) has achieved great success in intelligent agent control in recent years~\cite{arulkumaran2017deep, ze2023visual, zheng2024texttt}. 
It has wide applications in various domains such as autonomous driving, electronic sports, and robotic control \cite{liang2018cirl, kiran2021deep, lample2017playing, nair2018visual}.
Due to the high dimensionality of visual signals and the inefficiency of RL interactions, the algorithm struggles to understand the environment well, making it difficult to capture optimal rewards and lacking interpretability. 
Efficiently extracting task-relevant representations from visual observations is crucial for breaking through the bottleneck of RL.

\begin{figure}[h!]
\centering
\includegraphics[width=\linewidth]{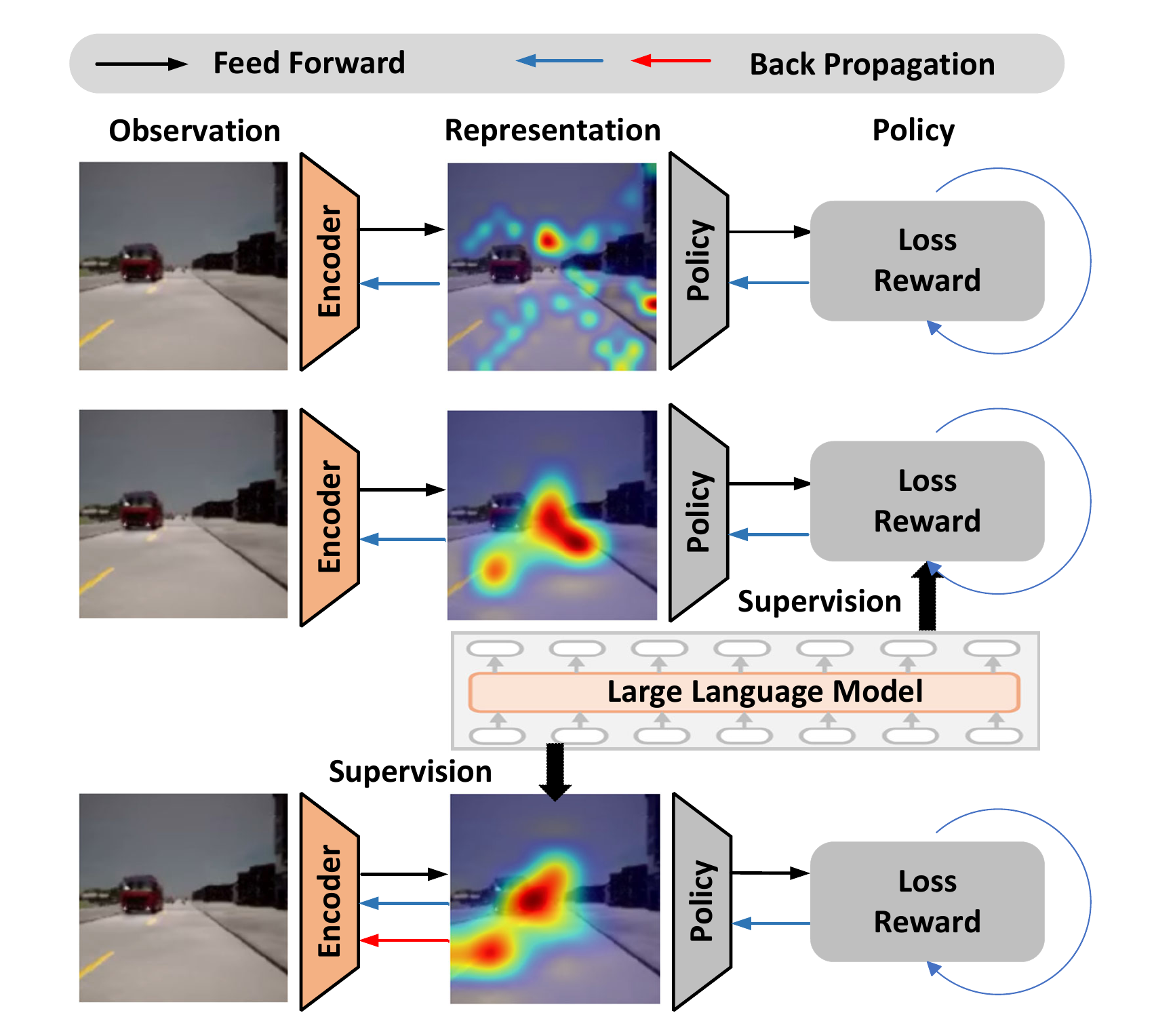}
\caption{
        (a) In the first row, the overly large sampling space of RL leads to difficulty in capturing key objectives in extreme scenarios;
        (b) In the second row, due to the complex high-dimensional feature space and the back propagation, guidance at the policy level cannot ensure that the encoder extracts reliable features;
        (c) In contrast, our method can fully take advantage of the capability of VLMs to enhance the task-specific representations. 
}
\label{fig:motivation}
\end{figure}

Previous researches leverage diverse state abstraction approaches including observation reconstruction~\cite{vemprala2021representation, yu2022mask},  transition dynamics prediction~\cite{gelada2019deepmdp} and bisimulation~\cite{zhang2020learning}, resulting in the issue of high cost and data redundancy~\cite{wang2024llm}. 
Consequently, a question arises regarding the existence of a more efficient method for explicitly extracting task-relevant representations.

Promisingly, Large Language Models (LLMs) have been actively developed in recent years, bridging human interaction and reasoning~\cite{wang2024llm, gbagbe2024bi, hu2024agentscomerge}. Based on the advancements in LLMs, they can provide a more holistic understanding of the environment, allowing agents to respond more effectively to various scenarios with human-like logic~\cite{han2024dme, huang2024understanding}. 
Some works leverage LLM to guide the learning of RL at policy level~\cite{chen2024vadv2, ma2024explorllm, hu2024agentscomerge} and indicate the enormous potential of LLMs.
Due to the long forward propagation chain of the RL model, the guidance at the policy level cannot effectively enhance the extraction capability of representations, particular in complex visual input tasks (as shown in Fig.~\ref{fig:motivation} (b)). 
This motivates the idea that the common-sense knowledge embedded in LLM can also be exploited to enhance the extraction capability of task-relevant representations at the feature level, which is shown in Fig.~\ref{fig:motivation} (c).

In order to address the aforementioned limitations, we introduce a novel Enhanced Semantic Motion Representation (Semore) Learning framework for visual RL, which employs a two-stream network to separately extract semantic and action representations. This design can decouple different features, thereby fully utilizing the LLM to guide the representation learning.
Specifically, the semantic stream models the environmental semantics and can identify key objects in the scenario, while the motion stream models motion clues from the residual frame of adjacent frames. 
To align the extracted representations and the actual surrounding environment, we introduce a VLM-based feature-level supervision module. 
We utilize VLM to generate task-specific feature masks, highlighting key regions in the observations.

Semantic and motion representations have strong complementarity and therefore can be enhanced through interaction with each other. 
Unlike previous work that used transformer-based networks to fuse two specific feature maps, we inject the feature map generated by the VLM during the training process.
Specifically, we adopt feature similarity loss to align the extracted semantic features to the VLM semantics for the semantic path. Meanwhile, we adopt cross-attention between motion features and VLM semantics to enhance the motion representation of key regions.
Essentially, the supervision for both pathways can let encoders to focus on key regions. Note the interaction between semantics and motion is spontaneously achieved, with the knowledge-aware features provided by the VLM serving as a mediator in this process.
Both semantic and motion representations are enhanced and fused for decision-making.

In summary, the contributions of this paper are three-fold:
\begin{itemize}
    \item We propose Semore, a novel VLM-based visual reinforcement learning framework that can enhance representations by integrating VLM-based common-sense knowledge guidance at the feature level.
    \item We designed a decoupled supervision module. For the semantic flow, we use explicit supervision for alignment, while for the motion flow, we use cross-attention to guide the focus areas. 
    \item We conduct comprehensive experiments using Carla benchmarks. Experimental results demonstrate the state-of-the-art performance of our proposed method and the effectiveness of the corresponding components.
\end{itemize}

\section{Related Works}
\label{sec:related}

\textbf{Visual Reinforcement Learning.} In vision-based RL, agents extract compact representations from low-dimensional visual observations to achieve decision-making.
In this process, representation learning is the key to improving the performance of visual RL.
Existing works can be roughly divided into three main approaches: (i) data augmentation technique \cite{huang2023simoun, zhang2020learning}; (ii) self-supervised representations \cite{castro2020scalable, hansen2020self}; (iii) modeling environment dynamics \cite{pan2022iso, fu2021learning, lee2020context}. 
However, due to the extensive exploration required in the RL process, existing methods without any prior knowledge struggle to efficiently extract representations, especially in complex environments.
The emergence of LLMs and VLMs brings new opportunities for addressing this issue.

\textbf{Dual-stream Network.} Dual-stream networks are particularly popular for extracting diverse representations \cite{gao2018motion, simonyan2014two, wang2024emiff}. 
Generally, this structure is used for encoding heterogeneous modalities such as point cloud and text for feature fusion~\cite{xiang2023hm, liu2023bevfusion, liang2022bevfusion}.
Some studies have demonstrated that separately extracting different representations containing specific information from images can achieve better performance than single-stream networks in extracting diverse representations~\cite{kim2021motion, liu2021emergence, huang2023video,liu2022learning}.
Simoun~\cite{huang2023simoun} adopts this design in visual RL learning and constructs a structure interaction module to leverage the correlations of the dual-stream features.
We adopt a two-stream structure to decouple the semantic and motion representation learning, allowing the VLM to separately supervise the feature extraction and interaction.

\textbf{VLM-based Learning.} 
VLMs have shown significant potential in learning high-quality representations for diverse downstream tasks~\cite{du2024lami, singh2022flava, chen2025asynchronous}. Their success largely stems from training transformer architectures on large-scale datasets of image-text pairs sourced from the web, using contrastive learning techniques. Notably, CLIP ~\cite{radford2021learning} proposed a promising alternative that directly learns transferable visual concepts from large-scale collected image-text pairs.
In this paper, we first use the VLM to retrieve semantic information from the observations, such as relevant objects, and then use a clip-based approach to generate the corresponding visual features, thereby embedding common-sense knowledge into the representation learning.

\section{Methodology}
\label{sec:method}

\begin{figure*}[!t]
    \centering
    \includegraphics[width=0.99\linewidth]{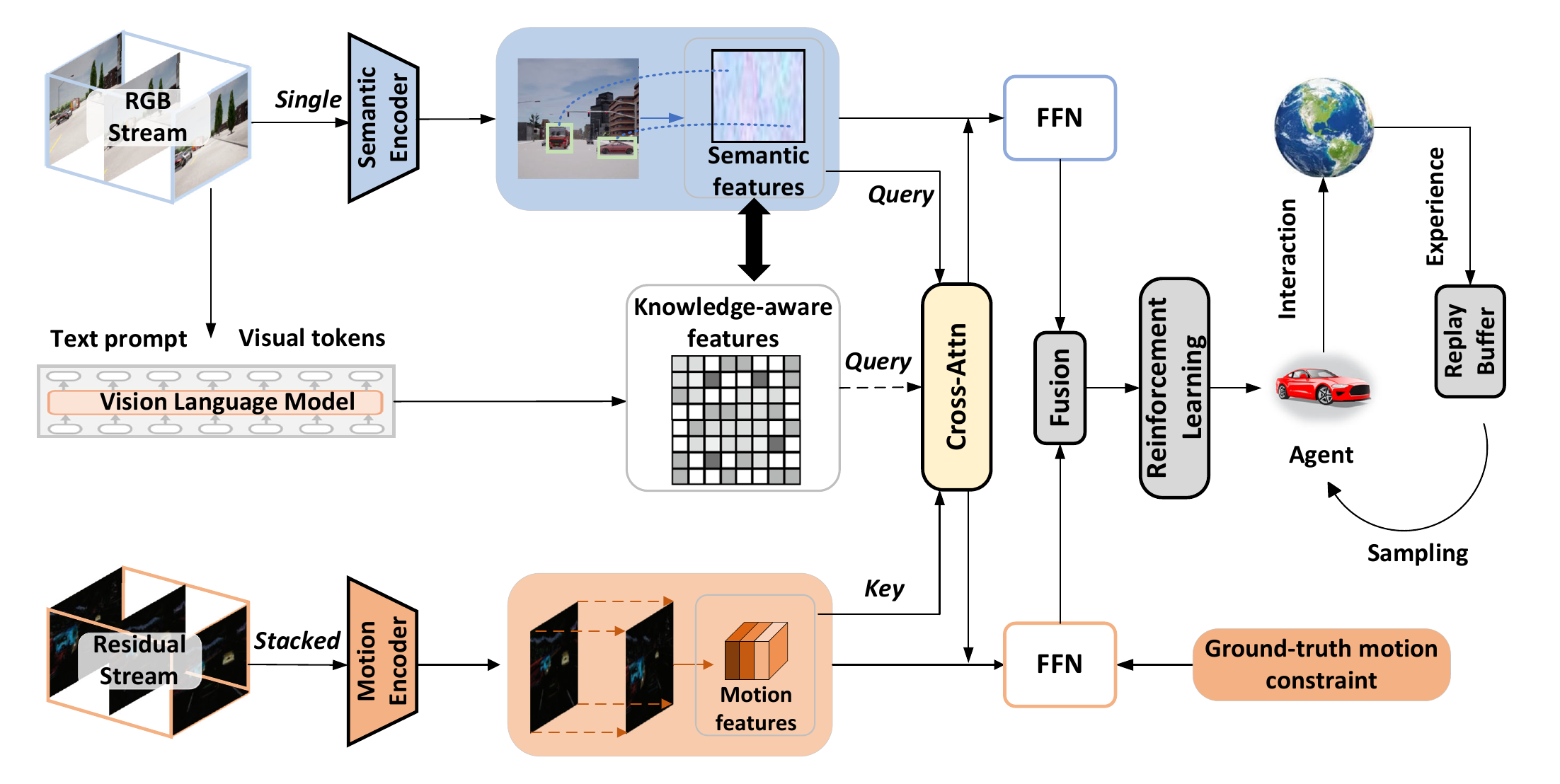}
    \caption{
     The overall VLM-guided learning framework. It integrates two key modules: 1) the VLM-guided semantics module employs the similarity loss to explicitly supervise the extraction of semantic representations, while the motion supervision module introduces knowledge-aware features into the motion extraction using bidirectional cross-attention.
    }
    \label{fig:framework}
\end{figure*}

The overall framework is illustrated in Fig.~\ref{fig:framework}.
We start by formalizing the task of visual RL and then discuss the details of Semore.

\subsection{Problem Formulation} 
\label{sec:problem}

Visual RL can be normally formulated as a Partially Observable Markov Decision Process (POMDP), denoted as a tuple $\mathcal{M} = <\mathcal{O}, \mathcal{S}, \mathcal{A}, \mathcal{P}, \mathcal{R}, \gamma >$, where $\mathcal{O}$ denotes the observation space containing RGB frames $o_t$ at different time step and $\mathcal{A}$ denotes the action space. 
The interaction process of the agent in a POMDP can be defined as follows: (i) the agent perceives visual observations $o_t$; (ii) the agent then selects an action $a_t \in \mathcal{A}$ based on a stochastic policy $\pi (a_t|o_t)$. $\mathcal{P}(o_t, a_t)$ is the observation transition, $\mathcal{R}(o_t, a_t)$ is the reward funciton, and $\gamma$ denotes the discount factor. The goal of this formulation is to find an optimal policy that maximizes the expected cumulative reward based on the visual observations across the entire traversal of MDPs:
\begin{equation}
    J(\pi) = \sum\limits_{t} E_{(o_t, a_t) \sim \pi}[\mathcal{R}(o_t, a_t)]
    \label{POMDP}
\end{equation}
During training, $\pi$ is used to interact with the environment and the generated experience is stored in a replay buffer $\mathcal{B}$. 

\subsection{Feature Extraction}
\label{sec:encoder}

\textbf{Semantic Encoding.} 
Semantic information is vital for visual representations as it can provide environmental understanding for the agent. Specifically speaking, the agent can identify objects or events relevant to its control based on visual representations from observations, thereby improving the decision-making capabilities.
Semantic encoding aims to extract the visual semantic representations of the surrounding environment from the observed frames. 
Semantic encoder adopts a four-layer convolutional structure with $3 \times 3$ kernel size and ReLU non-linearity. And the output feature map of the last convolution layer is represented as $F_{s}^{t} \in R^{C \times H \times W}$. Then we use a fully connected layer with layer normalization to reduce the dimension of $F_{s}^{t}$ and output compact feature vector $f_s^t$.


\textbf{Motion Encoding} 
Motion information is critical for vision-based agents to understand the dynamics of the surrounding scenarios. In other words, motion features enable the agent to have predictive capabilities, allowing it to make more reasonable decisions.
The goal of motion encoding is to extract the motion features at the pixel level (such as the movement of objects within the perception range) from multiple adjacent visual frames. 
Given a sequence of 3 adjacent observation frames $[o_{t-2}, o_{t-1}, o_t]$ sampled from the replay buffer, we can obtain the motion input by residual of adjacent frames $[o_{t-1}-o_{t-2}, o_t-o_{t-1}]$.
Similar to semantic encoding, the motion encoder adopts a four-layer convolutional encoder with a different number of first-layer's input channels to extract feature map $F_m^t \in R^{C \times H \times W}$.

\subsection{VLM-guided Semantics} 
\label{sec:vlm-exlipcit}

\begin{figure*}[!t]
    \centering
    \includegraphics[width=1.0\linewidth]{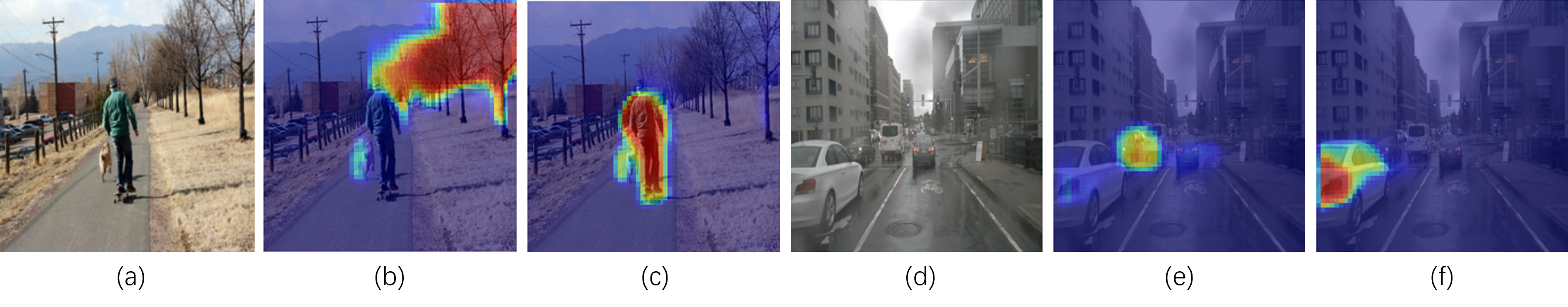}
    \caption{
    The illustration of VLM-generated knowledge-aware representations. (a) and (d) show the road scenarios of rural and urban areas, respectively. The text prompts are: (b): The right trees; (c): The pedestrian; (e): The white van on the left side of the black car ahead;
  (f): The left white car.
    }
    \label{fig:vlm generation}
\end{figure*}

In the absence of supervision on the feature level in previous visual RL models, the extracted representation of observations is not guaranteed to align with the expressive nature of the real environment. In particular, the long propagation chain and the vast sampling space of RL make it difficult to learn ground-truth representations through the mere supervision of the policy level. The difference from the true representation imposes inherent limitations on the model.
To mitigate this shortfall, we introduce VLM-guided representations by using the pretrained VLM and clip-based image segmentation model CRIS~\cite{wang2022cris}, as illustrated in Fig.~\ref{fig:vlm generation}. The core objective is to align the extracted representations with the ground-truth as possible, thereby enhancing the agent's understanding of the environment.

We first employ the pretrained VLM to extract comprehensive semantic information from raw visual observations.
The VLM takes a single frame $o$ as input and uses a visual encoder $g_V$ to extract the visual features, which are then converted into language embedding tokens. Meanwhile, the prompt $Pm$ is fed into the text encoder $g_T$ to obtain text tokens. This can be formulated as: 
\begin{equation}
\begin{aligned}
    H_V = g_V(O) \\ 
    H_T = g_T(Pm)
\end{aligned}
\end{equation}
where $H_V$, $H_T$ are the visual and text tokens, respectively. Then the visual tokens $H_V$ and the text tokens $H_T$ are fed into the VLM $f$ for generating responses:
\begin{equation}
    Y = f([H_V, H_T])
\end{equation}
where $Y$ is the output of task-specific prompts, as the text semantics. 

Given the text semantics, we need to generate a semantic feature map of the observations for representation learning. 
Although CLIP~\cite{radford2021learning} learns powerful image-level visual concepts by aligning the textual representation with the image-level representation, this type of knowledge is suboptimal for referring image segmentation, due to the lack of more fine-grained visual concepts.
Hence, we apply CRIS~\cite{wang2022cris}, a clip-driven image segmentation framework to accurately 
generate more discriminative visual representations through the alignment of text and visual features at the pixel-level. Specifically, given image $o$ and text sequence $Y = {y_1, ..., y_n}$, CRIS can compute and output a similarity map. Then we use the sigmoid function~\cite{cybenko1989approximation} to segment specific objects in the image, thereby generating corresponding high-confidence masks. We add the feature masks of different objects together to obtain the complete mask map.
This process can be described as:
\begin{equation}
\begin{aligned}
h_{y} &= CRIS(o, y), y \in Y \\
h_{ka} &= Sigmoid(h_{y}) \\
H_{ka} &= \sum_{y \in Y} h_{ka}
\end{aligned}
\end{equation}
where $o$, $Y$ is the input observed frame and text semantics. $Sigmoid$ denotes the Sigmoid function.


To impose effective supervision on semantics, we employ a similarity loss~\cite{zhao2016loss} between extracted and knowledge-aware representations, which is denoted as:
\begin{equation}
L_{S-G} =  \sum_{\mathcal{T}} \frac{||{F}_s(t) - \hat{H}_{ka}(t) ||_1}{||\hat{H}_{ka}(t)||_1}
\label{equa:simi}
\end{equation}
where $\hat{H}_{ka}$ is obtained by input $H_{ka}$ into the semantic encoder, thus having the same dimension as $F_{s}$.
Based on this process, VLM can leverage common-sense information to provide explicit guidance for semantics extraction. 
This guidance enables the visual encoder to focus on critical factors from the observations, thereby allowing it to detect objects that are key to control.

\subsection{Motion Enhancement and Interaction} 
\label{sec:interaction}


In previous work, popular approaches to handling different features were to use attention mechanisms for interaction, thereby outputting fused features. However, this process is a black-box operation, and supervision exists only at the end-point, thus lacking a comprehensive understanding. 
In particular, when processing motion and semantic features, attention-based fusion essentially aggregates them from different spaces into a common space. However, due to the impact of sampling efficiency, the learned fusion space is prone to overfitting and has low interpretability. To address this issue, we use VLM to enhance the motion features and guide the interaction between semantics.

\begin{figure}[htbp]
    \centering
    \includegraphics[width=0.5\textwidth]{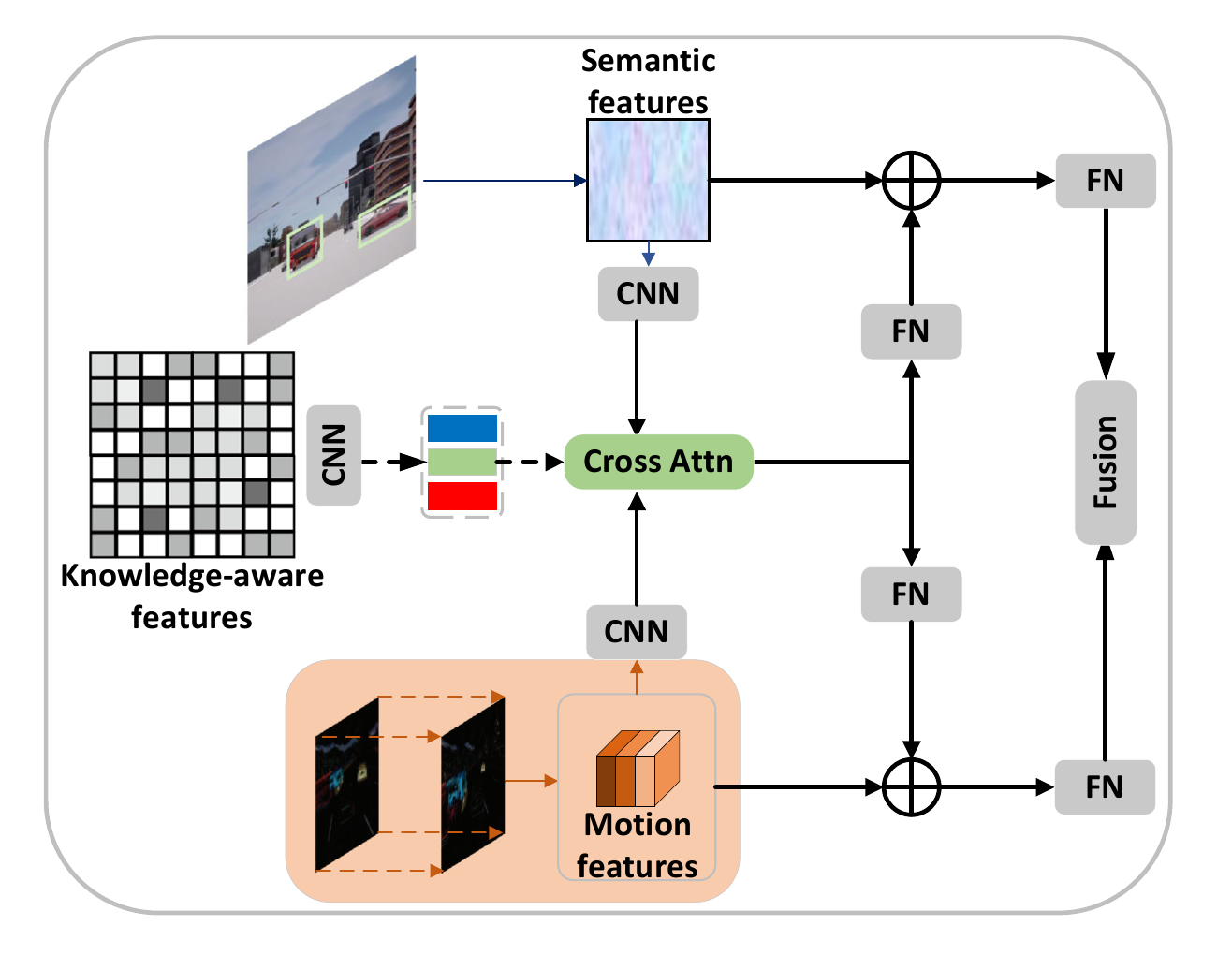} %
    \caption{Interaction of motion and semantic features. (Fusion is concat and knowledge-aware features are only used in training stage.).} %
    \label{fig:interaction} %
\end{figure}
Since motion information is extracted using frame differences, which is sparse. This makes it difficult to apply feature alignment for supervision. Therefore, during the training phase, we employ bidirectional cross-attention to guide the motion encoder in focusing on key areas using VLM-generated features. 
Specifically, as shown in Fig.~\ref{fig:interaction}, during the training, the input of the interaction module are knowledge-aware feature map ${H}_{ka}$ and a motion feature map $F_m$. Then an interactive attention map $X$ can be obtained: 
\begin{equation}
        X = \sigma(\sigma({\Tilde{H}_{ka}}^{\mathrm{T}}  \Tilde{F}_m) + \sigma( \Tilde{F}_m  {\Tilde{H}_{ka}}^{\mathrm{T}}  ) )
    \label{interaction1}
\end{equation}
where $\Tilde{H}_{ka}$ is obtained by inputting to a convolution layer for reducing the spatial complexity, $\Tilde{F}_m$ is as well. $\sigma$ denotes the Softmax function. 
Then we use the interactive feature map containing both semantic and motion information to simultaneously enhance original representations. Specifically, we separately use a fully connected layer to process the interaction and then add them to the feature maps of the semantic and motion feature maps:
\begin{equation}
\begin{aligned}
    F_s &= F_s + FN(X), \\
    F_m &= F_m + FN(X) 
\end{aligned}
\label{interaction2}
\end{equation}
where $FN$ denotes the fully connected layer. In this way, the semantic and motion representations complement each other, thereby enhancing understanding. Note that the semantics generated by VLM are only used to compute attention weights during training, while in reality, it is the semantic features and motion features that interact.

To ensure the motion encoder can effectively extract sufficient features and remove noise to prevent redundancy, we adopt a transition constraint via an MLP predictor $\mathcal{P}_m$. Specifically, the obtained feature vector $f_t^m$ and action $a_t$ at time step $t$ is input into the motion predictor. And the predictor can predict future features, thereby enhancing the encoder's ability to extract motion information. Then the transition loss can be defined as:
\begin{equation}
    \mathcal{L}_{trans} = ||\mathcal{P}_m(f_t^m,a_t), f_{t+1}^m||_2^2
\end{equation}
where $\mathcal{P}_m$ represents the motion predictor and $||||_2$ is the L2-norm.

After completing the interaction, we use fully connected layers to reduce the dimensions of $F^s_t$ and $F^m_t$, obtaining compact features $f^s_t$ and $f^m_t$. Then we concatenate them and the final representation is $f_t = [f_t^s, f_t^m]$. 
To reduce noise, we use a prediction head to further purify the information related to RL rewards. Motivated by DeepMDP~\cite{gelada2019deepmdp}, we utilize a reward predictive head by incorporating the tractable reward and state head from DeepMDP~\cite{gelada2019deepmdp} to predict the reward value of each observation-action pair: 
\begin{equation}
    \mathcal{L}_R = ||\mathcal{R}(f_t, a_t)-r_{t+1}||
\end{equation}
where $r_{t+1}$ is the actual reward value at the next time step, which is returned from the interaction with the environment.

\subsection{Reinforcement Learning based on Semore}
\label{sec:RLleanrning}

We adopt the baseline RL algorithm SAC~\cite{haarnoja2018soft} to maximize the expected cumulative reward to find the optimal policy via approximating the action-value $Q$ and a stochastic policy $\pi$ based on an $\alpha$-discounted maximum entropy $\mathcal{H}(\cdot)$.
The action-value function $Q$ is learned by minimizing the soft Bellman error and the soft state value $V$ can be estimated by sampling an action under the current policy. The above process can be formulated as:
\begin{equation}
    \begin{aligned}
        J(\pi) &= \sum_{t} E_{(o_t,a_t) \sim \pi} [r(o_t,a_t)+\alpha \mathcal{H}(\pi(\cdot | o_t))], \\
        \mathcal{L}_Q &= E_{(o_t,a_t)} (Q(o_t,a_t)-(r_t+\lambda V(o_{t+1})))^2, \\
        V(o_{t+1}) &= E_{a_{t+1} \sim \pi} [\Tilde{Q}(o_{t+1},a_{t+1})-\alpha log \pi (a_{t+1}|o_{t+1})],
    \end{aligned}
\end{equation}
where $\Tilde{Q}$ denotes the exponential moving average of the parameters of $Q$. And the policy is optimized by decreasing the difference between the exponential of the soft-Q function and the policy:
\begin{equation}
    \mathcal{L}_{\pi} = E_{a_t \sim \pi}[\alpha log \pi (a_t|o_t)-Q(o_t,a_t)].
\end{equation}

\subsection{Training Details}
\label{sec:training}

\begin{figure}[htbp]
    \centering
    \includegraphics[width=0.48\textwidth]{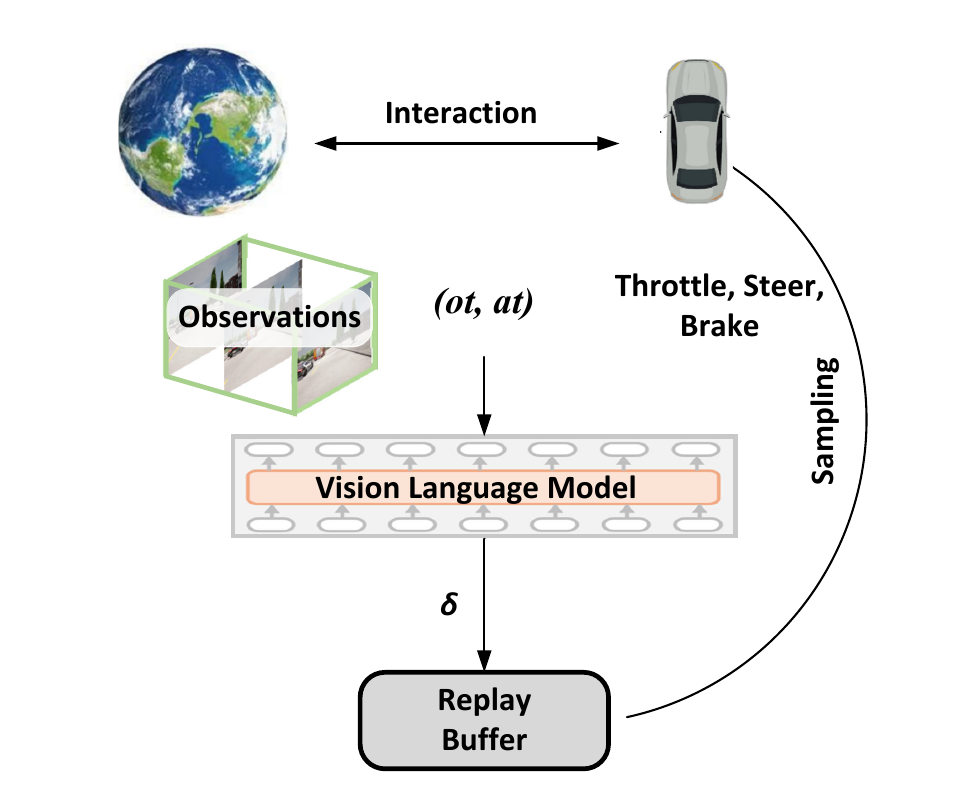} %
    \caption{VLM-based selective replay buffer.} %
    \label{fig:replay} %
\end{figure}
\textbf{Selective Replay Buffer} RL agents are very likely to perform ineffective exploration in the initial stage due to the lack of prior knowledge.
To alleviate the low exploration efficiency for large continuous action space in visual RL that often prohibits the use of challenging tasks, we design the selective replay buffer to provide better exploration.
The intuition is that if the agent is provided with some positive training data such as expert supervision at the beginning, it can acquire a certain level of initial execution capability, thus avoiding the high cost of excessive random sampling.
Specifically, our RL framework will be first warmed up by learning knowledge via the observation-action pairs that are deemed qualified by the LLM to initialize the action exploration in a reasonable space. 

LLM is not able to make precise decision signals, but can offer macro-level guidance, such as braking, turning left, or turning right. Therefore, it can be fully leveraged to evaluate the reasonableness of observation-action pairs.
As illustrated in Fig.~\ref{fig:replay}, the generated pair $(o_t, a_t)$ is fed into the image LLM and we can obtain the output that is reasonable or unreasonable.
We additionally adopted a decay factor $\delta$ to represent the probability of adding the observation-action pair to the replay buffer. Its initial value is set to 1, and it decreases along the training process. When its value reaches 0.5, it indicates that no further selective additions will be made, and instead, all interactions with the environment will be added to the replay buffer.

\textbf{Training Objective.} 
Based on the fused semantic motion representations, Semore learns from visual to control signals in an end-to-end manner via optimizing the following equation:
\begin{equation}
    \mathcal{L} = \underbrace{\mathcal{L}_{trans}}_{motion} + \underbrace{\mathcal{L}_{S-G}}_{semantic} + \underbrace{\mathcal{L}_{R}}_{state}  + \underbrace{\mathcal{L}_{\pi} + \mathcal{L}_{Q}}_{RL}
\end{equation}
where the objective jointly considers the semantic and motion representations, as well as the purification of reward-related information for RL learning.

\section{Experiments}
\label{experiments}

\subsection{Experimental Setup}
\label{experimental setup} 
\begin{figure}[htbp]
    \centering
    \includegraphics[width=0.45\textwidth]{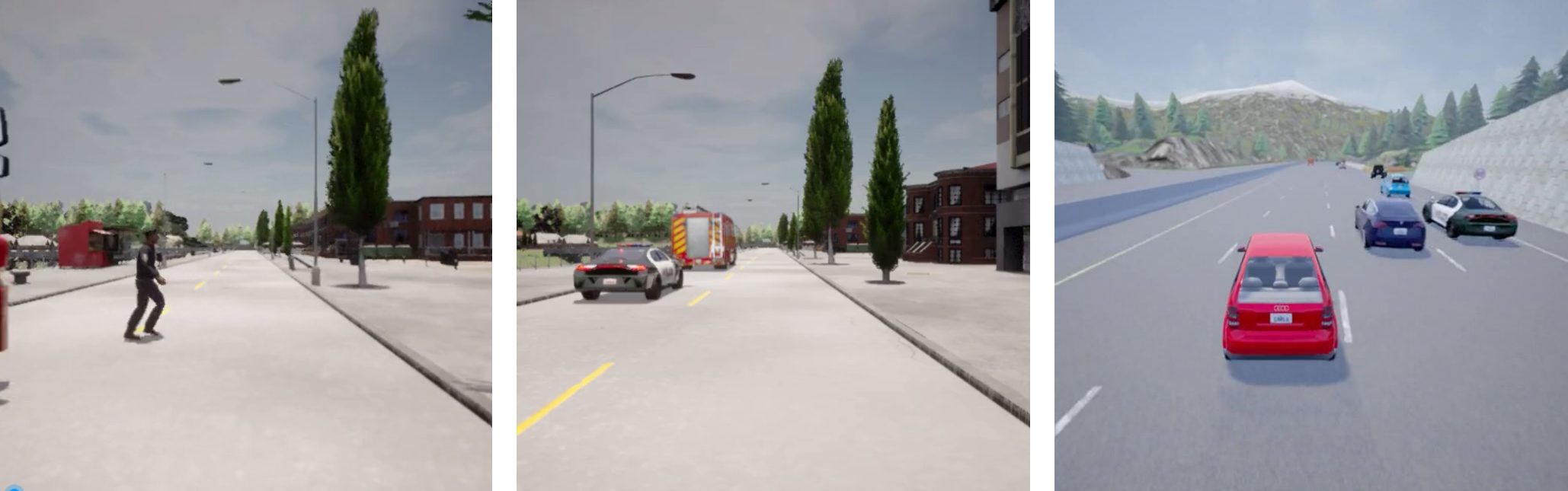} %
    \caption{Visulization of the CARLA scenarios, where the left column is JW, the middle column is HB, and the right column is HW.} %
    \label{fig:scenario} %
\end{figure}

To evaluate our approach under realistic and challenging vision-based environments,
we employ the CARLA simulator~\cite{dosovitskiy2017carla}, which is a widely used open-source simulator for autonomous driving~\cite{liang2018cirl, xu2024dmr, huang2023hierarchical}. 
CARLA provides a rich and realistic urban environment to evaluate autonomous driving agents in various traffic scenarios. As shown in Fig.~\ref{fig:scenario}, we evaluate our method in three traffic scenarios: the HighBeam (HB) scenario, where the ego-vehicle encounters a cyclist, JayWalk (JW) scenario, where the ego-vehicle encounters both stationary and moving pedestrians intermediately and HighWay (HW) scenario, where the ego-vehicle is driving on an eight-lane highway with numerous vehicles traveling in the same direction. Similar to \cite{zhang2020learning, fan2021secant}, the reward function can encourage the agent to avoid crashes with other moving and static objects and travel as long as possible.  
We set the single camera on the ego-vehicle's roof with a view of 60-degree.

Our method is implemented based on SAC~\cite{haarnoja2018soft} and DeepMDP~\cite{gelada2019deepmdp}. The same encoder network architecture and training hyperparameters are adopted for all comparative methods.
The spatial resolution of the input RGB images is $128 \times 128 \times 3$. All methods are trained for 110k frames using 5 random seeds to report the mean and standard deviation of the rewards. And more details can be found in the attached supplementary material.

\begin{figure}[htbp]
    \includegraphics[width=0.47\textwidth]{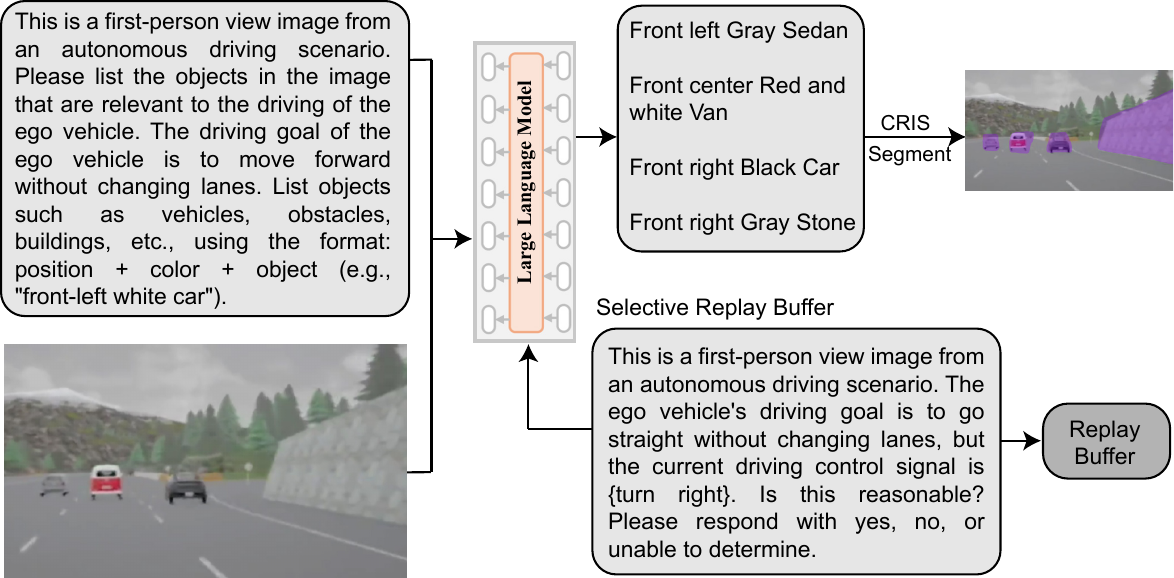} %
    \caption{The input prompts for the VLM.} %
    \label{fig:visual} %
\end{figure}

We adopt Qwen2-VL-7B-Instruct~\cite{wang2024qwen2} in the experiments and Fig.~\ref{fig:visual} shows the input prompts for the VLM.

\textbf{Methods Compared:} 
We consider the following baseline methods for comparison: 1) SAC~\cite{haarnoja2018soft}, a widely-used RL algorithm based on $\alpha$-discounted maximum entropy; 2) Flare~\cite{shang2021reinforcement}, a multi-frame visual RL method that utilizes temporal information through latent vector differences. 3) CURL~\cite{laskin2020curl}, which integrates contrastive learning with model-free RL with minimal changes to the architecture and training pipeline. 4) DrQ~\cite{yarats2021image}, built upon the SAC by adding a convolutional encoder and data augmentation in the form of random shifts.
5) DeepMDP~\cite{gelada2019deepmdp}, a latent model of an MDP and has been trained to minimize two tractable losses: predicting the rewards and predicting the distribution of the next latent states.
6) Simoun~\cite{huang2023simoun}, a dual-stream visual RL method that simultaneously extracts appearance and motion information, and enhances representations through interaction and intrinsic rewards.

\begin{table*}[!t]
\centering
\footnotesize
\resizebox{1.0\linewidth}{!}{
\begin{tabular}{c|c|c|c|c|c|c|c|c} 
\toprule
Scen. & Metrics & SAC & CURL & Flare & DrQ & DeepMDP & Simoun & Ours  \\
\midrule
\multirow{5}{*}{JW} &  Episode reward $\uparrow$ &  103$\pm$72  &  112$\pm$81   &  136$\pm$52  &  97$\pm$59   &  146$\pm$48   &  168$\pm$79  &  201$\pm$54  \\
&  Distance (m)  $\uparrow$  &  143$\pm$58 &  128$\pm$61  &  123$\pm$68  &  109$\pm$39   & 132$\pm$53    &  208$\pm$74  & 233$\pm$66    \\ 
& Crash intensity $\downarrow$  &   4633$\pm$184    &  3829$\pm$153   &   2974$\pm$171   &  3013$\pm$187   &  2627$\pm$98   &   2382$\pm$103    & 2043$\pm$98   \\ 
& Average steer (\%) $\downarrow$  &  15.40 & 14.79  & 12.53  &  13.29 &  10.88 & 13.69  &  13.91  \\ 
& Average brake (\%) $\downarrow$  & 2.06 & 2.83 & 2.27 & 2.02 & 1.95 & 2.80 & 2.86   \\ 
\midrule
\multirow{5}{*}{HB} & Episode reward $\uparrow$ & 73$\pm$62 & 86$\pm$67 & 82$\pm$54 & 93$\pm$71 &  101$\pm$45 &  104$\pm$66  & 123$\pm$60 \\ 
&  Distance (m) $\uparrow$ &  86$\pm$43 & 96$\pm$57 & 92$\pm$56 &  91$\pm$64 & 112$\pm$59  &  128$\pm$61    &  163$\pm$52  \\ 
&   Crash intensity  $\downarrow$  &  4850$\pm$153 & 4145$\pm$193 &  3692$\pm$133   & 3511$\pm$176  &  2921$\pm$102  &  2604$\pm$109   & 2418$\pm$103 \\ 
& Average steer (\%) $\downarrow$  &  15.53  & 14.20  & 12.13 &  12.97  &  10.14 &  13.56 & 13.03  \\ 
& Average brake (\%) $\downarrow$  &  2.43  & 3.11 &  2.94  &  2.81  & 2.57  &  3.06  &  3.08  \\ 
\midrule
\multirow{5}{*}{HW}   &  Episode reward $\uparrow$ & 126$\pm$24 & 134$\pm$19 &  138$\pm$29  &  167$\pm$23  &  182$\pm$35   &   268$\pm$30    &  316$\pm$33   \\ 
&  Distance (m) $\uparrow$  &  102$\pm$17  &  137$\pm$35   &  114$\pm$22   &  108$\pm$29    &  129$\pm$22  &  202$\pm$22  &  263$\pm$17  \\ 
&   Crash intensity  $\downarrow$  &  3870$\pm$98  &  3122$\pm$104  &  2548$\pm$91   &  2487$\pm$98   & 2153$\pm$76  &  1816$\pm$63  & 1671$\pm$56  \\ 
& Average steer (\%) $\downarrow$  &   17.03 &  15.89   & 12.28  &  15.95 & 10.56 & 15.10  &  14.24  \\ 
& Average brake (\%) $\downarrow$  &  1.84  & 2.56 &  1.82  &  1.63  &  1.54 &  2.14 &  2.13  \\ 
\bottomrule
\end{tabular}}
\caption{
Quantitative results of different models for driving policies. 
( $\uparrow$ indicates that larger is better while $\downarrow$ means opposite. ) 
}
\label{tab:overall}
\end{table*}

\subsection{Overall Performance} 

\begin{figure}[htbp]
    \centering
    \includegraphics[width=0.4\textwidth]{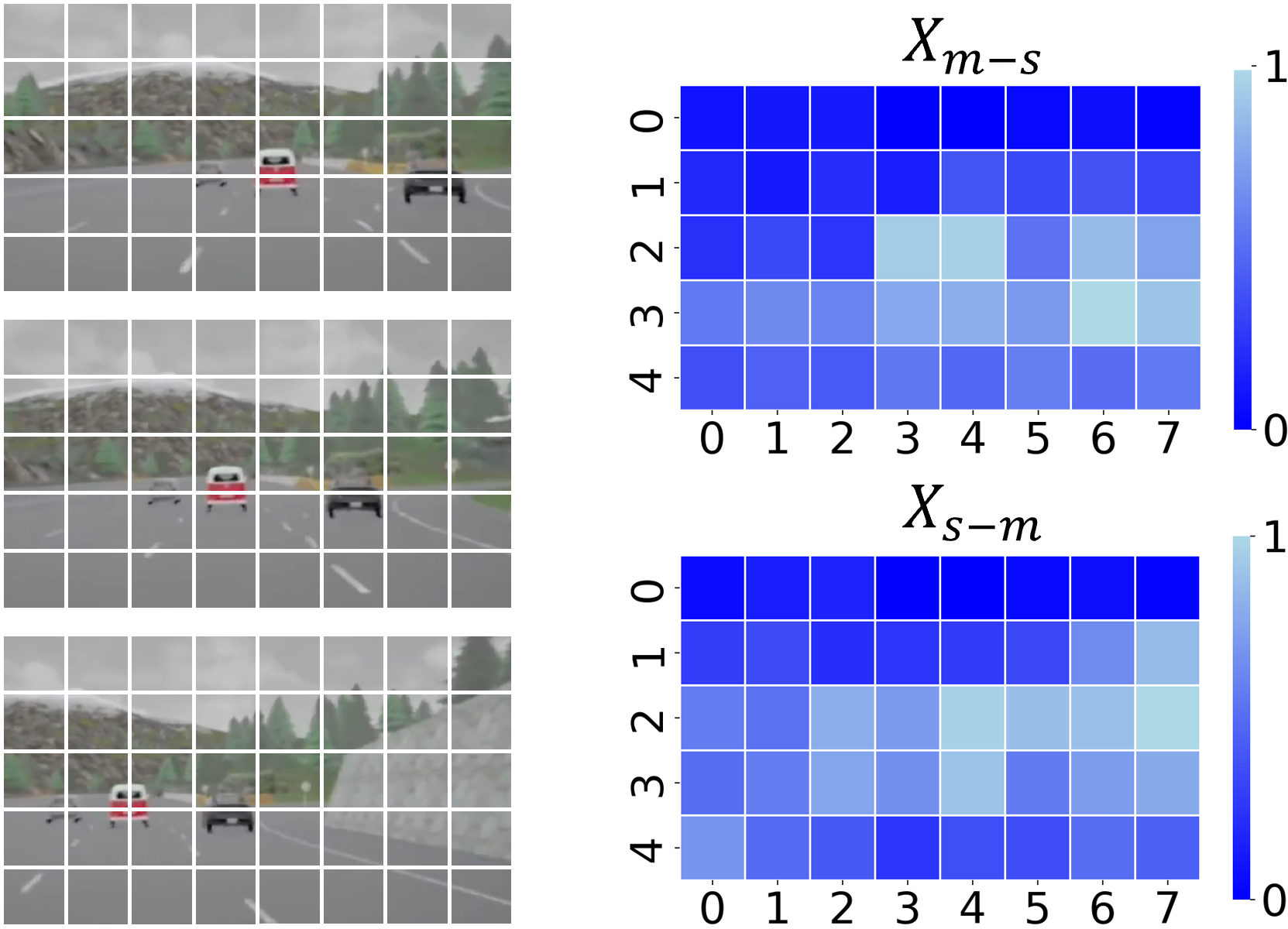} %
    \caption{
    Visualization of the feature masks.
    } %
    \label{fig:visual2} %
\end{figure}

We compare Semore with benchmarks and the results are shown in Tab.~\ref{tab:overall}. It can be observed that our method outperforms all other methods in terms of the episode reward. 
And the average driving distance is farther than other methods and the average crash intensity is lower.
In particular, the observed improvements over Simoun emphasize the effectiveness of VLM in guiding representation learning.
Note that our method did not achieve the best driving smoothness with a higher average brake and steer value. Combining the driving distance and crash intensity, this is likely because the comparative methods did not make appropriate obstacle avoidance decisions based on specific objectives. This is particularly evident in scenarios with higher traffic density. When the number of objects on the road increases, the ego-vehicle must take action to alter its current state for obstacle avoidance. Fig.~\ref{fig:visual2} visualizes the computed interaction feature masks using the Equation.~\ref{interaction2} of the HW scenario.

\vspace{-2mm}

\subsection{Ablation Study}

\textbf{Effectiveness of Semore Components.} 
To validate the contribution of each component, we incrementally incorporate individual components of the framework and obtain a series of models labeled M1 to M4. Specifically, M1 utilizes solely the semantic-stream branch for decision-making without the supervision of explicit similarity loss;
In M2, the semantic and motion branches are employed, and the features from both streams are directly concatenated to feed into the policy learning.
Both M3 and M4 utilize VLM-generated features as explicit supervision of semantic representation.
M3 leverages semantic supervision to align to the knowledge-aware representations in terms of Eq. \ref{equa:simi}.
M4 builds upon M3 by incorporating motion supervision and interaction.

Tab.~\ref{tab:components} shows the performance of each model. 
It is clear that M1 degrades to a conventional multi-frame input visual RL, while M2, by decoupling semantic and motion information, enhances the representation extraction capability.
This demonstrates the effectiveness of dual-stream design.
With the guidance of the VLM, M3 achieves significant performance improvements. However, due to its inability to effectively integrate motion information, and considering the highly dynamic nature of the scenes, its obstacle avoidance capability improves only marginally compared to M2.

\begin{table}[!htbp]
\centering        
\footnotesize
\begin{tabular}{c|c|c|c|c|c} 
    \toprule 
    \multicolumn{2}{c|}{\diagbox{Metrics}{Model}} & M1 & M2 & M3 & M4 \\
    \midrule
    \multirow{2}{*}{JW}  &  Distance(m)  &  127 & 163  & 174  &  233 \\ 
                     &  Intensity   &  3420  &  2582   &  2471  &  2043 \\ 
\midrule
    \multirow{2}{*}{HB}  &   Distance(m)  &  76  &  119   &   137    &    163    \\ 
      & Intensity  &  3846  &   2673  &    2549    &      2418      \\ 
\midrule
    \multirow{2}{*}{HW}  &  Distance(m)   &   146   &     203     &    238   &   263    \\ 
                       &  Intensity   &  3073  &  1940     &  1886    &   1671     \\ 
    \bottomrule
 \end{tabular}
 \caption{
Effect of components in Semore. 
} 
 \label{tab:components}
\end{table}
\vspace{-2mm}

\section{Conclusion}
In this paper, we propose Semore, a novel framework aimed at addressing the issue of limited representation learning capability in visual RL.
Semore can leverage knowledge-aware supervision in both semantic and motion representation learning under the guidance of VLM. 
Based on the decoupled two-stream network architecture, semantic extraction can be enhanced through feature alignment under explicit supervision.
Simultaneously, a bidirectional cross-attention mechanism is used to enhance motion extraction while achieving semantic-motion interaction.
Thus, the knowledge of the VLM is distilled into our encoders, thereby enhancing the representations.
Extensive experiments in different challenging scenarios demonstrate the efficacy and superiority.

\section{Acknowledgments}
This work is supported by Xiongan AI Institute, Didi Chuxing and Wuxi Research Institute of Applied Technologies, Tsinghua University under Grant
20242001120.

\bibliography{aaai2026}

\end{document}